\documentclass[11pt,a4paper]{article}
\usepackage[margin=1in]{geometry}
\usepackage{amsmath,amssymb,amsfonts}
\usepackage{booktabs}
\usepackage{graphicx}
\usepackage{float}
\usepackage{caption}
\usepackage{hyperref}
\usepackage{microtype}

\graphicspath{{paper_figures/}}
\title{\textbf{Conformalized Tree-Based Gradient Ensembles for Out-of-Distribution Detection and Plausible Recourse}}

\author{
  Mariam Zakaria Moussa Ali \\
  Department of Computer Science (AI \& Data Science), Arab Open University, Egypt \\
  \texttt{mariamza200400@gmail.com}
}

\date{}

\begin{document}
\maketitle

\begin{abstract}
Counterfactual recourse provides actionable feature adjustments that reverse unfavourable predictions made by automated decision models. However, when recourse search optimizes purely against classifier confidence or narrow proxy scores over gradient-boosted decision trees (\textsc{GBDT}), search procedures can exploit boundary artifacts and low-density regions---an instance of Goodhart's law---yielding out-of-distribution (OOD) recommendations that appear favourable nominally while lacking manifold plausibility. In this paper, we study conformalized OOD gating coupled with actionability-constrained counterfactual search across two tabular benchmarks: \textsc{Adult} and \textsc{Give Me Some Credit (GMSC)}. Calibrating finite-sample anomaly gates (\texttt{gate\_IF}, \texttt{gate\_LOF}, and the conjunctive \texttt{gate\_ALL}) at significance level $\delta = 0.05$ over isolation forest, local outlier factor, and $k$-nearest-neighbor signals, we audit generated counterfactuals against held-out \textsc{GMM} and \textsc{OCSVM} detectors alongside ablation studies (\texttt{gen10}, \texttt{gen100}, and \texttt{relax}). Across matched recourse pairs where both constrained and unconstrained searches succeed, \texttt{gate\_ALL} significantly reduces held-out \textsc{OCSVM} flags on \textsc{Adult} ($N=541$, $\Delta \text{flag}_{\text{OCSVM}} = -0.0296$, $p_{\text{Holm}} = 4.0\times 10^{-4}$) and lowers both held-out flags on \textsc{GMSC} ($N=540$, $\Delta \text{flag}_{\text{OCSVM}} = -0.0346$, $p = 1.0751\times 10^{-4}$; $\Delta \text{flag}_{\text{GMM}} = -0.1663$, $p = 3.5679\times 10^{-16}$) while increasing matched recourse cost ($\Delta \text{cost} = +0.3035$, $p = 1.3365\times 10^{-5}$) and narrowing overall feasibility. Our findings formalize the cost--feasibility--plausibility trade-off and demonstrate how conformal OOD gating audits recourse reliability on tabular ensembles.
\end{abstract}

\section{Introduction}
Machine learning models deployed in credit underwriting, lending, and socio-economic screening increasingly affect consequential human decisions. Beyond predictive accuracy, stakeholders require post-hoc explanations that are \emph{actionable}: counterfactual recourse identifies concrete changes to mutable features that move an unfavourable prediction toward a favourable threshold~\cite{uncertainty_ce,wachter2017counterfactual,ustun2019actionable}.

A central failure mode in counterfactual recourse arises from optimizing purely against a model score or heuristic proxy, which can push candidate states into sparse or out-of-distribution (OOD) regions of feature space~\cite{medix,tood}. Such counterfactuals may achieve high nominal classification probability while being implausible, fragile, or detached from the underlying data manifold~\cite{cp_ood,poyiadzi2020face}.

To examine reliability and plausibility in tabular decision support, three lines of research inform our study:
\begin{itemize}
    \item \textbf{Uncertainty and Plausibility in Recourse:} Evaluating predictive reliability and out-of-distribution shifts helps distinguish genuine actionable transitions from sparse or implausible states~\cite{uncertainty_ce,wachter2017counterfactual,ustun2019actionable}.
    \item \textbf{Structural, Attribution, and Tree-Based OOD Signals:} Beyond raw confidence scores, neuron-level relevance attribution (\textsc{NERO})~\cite{nero}, median gradient statistics under unlabeled wild-data contamination (\textsc{Medix})~\cite{medix}, and tree-embedding pairwise Hamming distances (\textsc{TOOD})~\cite{tood} capture distributional shifts across representations and ensemble partitions.
    \item \textbf{Conformal Calibration:} Conformal prediction provides distribution-free, finite-sample threshold calibration with marginal coverage $1-\delta$, mitigating optimistic threshold or metric estimates on finite calibration splits~\cite{cp_ood,vovk2005algorithmic,shafer2008tutorial,angelopoulos2023gentle}.
\end{itemize}

Furthermore, treating a surrogate score or single proxy as an optimization objective without distributional verification reflects \emph{Goodhart's law}~\cite{goodhart1975problems} and motivates manifold-aware plausibility checks~\cite{poyiadzi2020face}. This paper investigates conformalized OOD gating for constrained counterfactual recourse on \textsc{GBDT} models across \textsc{Adult}~\cite{kohavi1996adult} and \textsc{GMSC}~\cite{gmsc2011kaggle}, auditing plausibility transfer to held-out \textsc{GMM} and \textsc{OCSVM} detectors, performing paired Wilcoxon signed-rank tests over matched recourse pairs, and analyzing search-budget and constraint-relaxation ablations (\texttt{gen10}, \texttt{gen100}, and \texttt{relax}).

\section{Related Work}
\subsection{Counterfactual Recourse and Plausibility}
Foundational work on counterfactual explanations and actionable recourse established minimal intervention principles and actionability constraints for unfavourable decisions~\cite{wachter2017counterfactual,ustun2019actionable}. Subsequent studies highlighted that unconstrained search can yield off-manifold or unrealistic states, motivating plausibility and data-manifold constraints~\cite{poyiadzi2020face} alongside uncertainty and OOD estimation for counterfactual explanations~\cite{uncertainty_ce}. When search optimizes a single score without plausibility checks, Goodhart's law~\cite{goodhart1975problems} can produce high-scoring yet implausible recommendations.

\subsection{Out-of-Distribution Detection and Structural Signals}
Post-hoc OOD detection aims to identify inputs that diverge from the training distribution. Recent methods explore neuron-level relevance (\textsc{NERO})~\cite{nero}, median gradient statistics under wild-data contamination (\textsc{Medix})~\cite{medix}, and tree-ensemble partition geometry via average pairwise Hamming distance (\textsc{TOOD})~\cite{tood}. For tabular tree ensembles, density and neighborhood anomaly scorers (\textsc{Isolation Forest}, \textsc{LOF}, $k$-\textsc{NN}) and generative/margin models (\textsc{GMM}, \textsc{OCSVM}) provide complementary views of feature-space support.

\subsection{Conformal Prediction and Statistical Calibration}
Conformal prediction provides finite-sample calibration and distribution-free marginal validity guarantees under exchangeability~\cite{vovk2005algorithmic,shafer2008tutorial,angelopoulos2023gentle}. Connecting conformal calibration with OOD detection addresses finite-sample variability in anomaly thresholds and evaluation metrics at significance level $\delta$~\cite{cp_ood}.

\section{Methodology}
\subsection{Actionability-Constrained Recourse Search}
Let $x \in \mathcal{X}$ denote an unfavourable tabular instance scored by a trained \textsc{GBDT} classifier $f_\theta(x)$ with favourable class probability $p_\theta(y=1 \mid x)$. We restrict candidate modifications to an actionable domain $\mathcal{A}(x)$ enforcing domain rules:
\begin{itemize}
    \item \textbf{Immutability:} Non-actionable demographic and historical attributes remain fixed~\cite{ustun2019actionable}.
    \item \textbf{Monotonicity and Spike Protection:} Ordered features constrained to non-decreasing shifts respect directionality, and zero-inflated or high-mode-share attributes (such as \texttt{capital-loss} with mode share $0.953$ and \texttt{capital-gain} with mode share $0.917$ on \textsc{Adult}) are guarded against spurious jumps.
\end{itemize}
An evolutionary search optimizes candidates $x' \in \mathcal{A}(x)$ to cross a target favourable probability threshold $\tau_p$ while tracking normalized recourse \texttt{cost} (cumulative feature displacement scaled by training standard deviation), count of modified features (\texttt{n\_chg}), and local perturbation stability (\texttt{robust}, measuring prediction consistency under small input perturbations).

\subsection{Conformalized OOD Gates and Conjunctive Filtering}
Let $s_m(x)$ be the anomaly score from base detector $m \in \{\text{IF}, \text{LOF}, \text{kNN}\}$ fitted on training data. Using held-out calibration scores $\{s_m(x_i^{\text{cal}})\}_{i=1}^{n_{\text{cal}}}$, we compute empirical conformal quantile thresholds $\hat{q}_m$~\cite{shafer2008tutorial,angelopoulos2023gentle,cp_ood} at significance level $\delta = 0.05$ so that an instance passes gate $G_m$ when $s_m(x) \le \hat{q}_m$:
\begin{align}
G_{\text{IF}}(x) &= \mathbb{I}\!\left(s_{\text{IF}}(x) \le \hat{q}_{\text{IF}}\right), \quad
G_{\text{LOF}}(x) &= \mathbb{I}\!\left(s_{\text{LOF}}(x) \le \hat{q}_{\text{LOF}}\right), \\
G_{\text{ALL}}(x) &= \prod_{m \in \{\text{IF}, \text{LOF}, \text{kNN}\}} \mathbb{I}\!\left(s_m(x) \le \hat{q}_m\right).
\end{align}
Under exchangeability, each individual conformal threshold provides marginal false-alarm control at level $\delta$ on in-distribution calibration points, without guaranteeing joint coverage across all three detectors simultaneously or universal OOD detection. Empirically, single-detector gates (\texttt{gate\_IF}, \texttt{gate\_LOF}) guard against individual anomaly modes, whereas the conjunctive filter \texttt{gate\_ALL} enforces simultaneous agreement across all three calibrated scorers.

\subsection{Held-Out Plausibility Auditing}
To prevent circularity (``teaching to the gate''), we audit generated counterfactuals using held-out \textsc{GMM} and \textsc{OCSVM} detectors calibrated at the same $\delta = 0.05$ level on the calibration split and excluded from search-time gating. Reductions in held-out detector flags indicate plausibility transfer across distinct anomaly formulations~\cite{uncertainty_ce,poyiadzi2020face,cp_ood}.

\section{Experimental Setup}
We evaluate on two tabular benchmarks: \textsc{Adult}~\cite{kohavi1996adult} ($4{,}000$ evaluation queries across $5$ random seeds $\times$ $200$ unfavourable test instances per seed for each of the $4$ gate configurations at target probability $\tau_p = 0.6$, plus $1{,}800$ evaluation queries across $3$ ablation variants) and \textsc{Give Me Some Credit (GMSC)}~\cite{gmsc2011kaggle} ($6{,}160$ evaluation queries in the main gate benchmark). On \textsc{Adult}, the underlying \textsc{GBDT} classifier is evaluated across disjoint training, calibration-fit, calibration, and test partitions, achieving test $\text{Accuracy} = 0.8631$, $\text{Balanced Accuracy} = 0.7926$, $\text{AUROC} = 0.9163$, and Expected Calibration Error $\text{ECE} = 0.0351$.

We compare unconstrained recourse search (\texttt{none}) against \texttt{gate\_IF}, \texttt{gate\_LOF}, and \texttt{gate\_ALL}. For each configuration, we report:
\begin{itemize}
    \item \textbf{Feasibility and Conditional Recourse Metrics:} Feasibility rate (\texttt{found}, fraction of unfavourable queries achieving valid recourse) and conditional means over successful counterfactuals (\texttt{found} $= 1$) for achieved probability (\texttt{prob}), normalized \texttt{cost}, modified feature count (\texttt{n\_chg}), perturbation robustness (\texttt{robust}), and calibrated OOD flag rates (\texttt{flag\_IF}, \texttt{flag\_LOF}, \texttt{flag\_kNN}, \texttt{flag\_GMM}, and \texttt{flag\_OCSVM}). On \textsc{Adult}, empirical flag rates on real favourable test rows serve as a natural in-distribution reference baseline (\texttt{IF} $= 0.071$, \texttt{LOF} $= 0.044$, \texttt{kNN} $= 0.064$, \texttt{GMM} $= 0.079$, \texttt{OCSVM} $= 0.074$).
    \item \textbf{Matched-Pair Statistical Comparisons:} Because gating alters which queries succeed, direct differences between conditional table averages can reflect selection effects. We therefore evaluate paired differences ($\Delta = \text{gate} - \text{none}$) strictly on the intersection of queries where both methods succeed (\emph{matched pairs}: $N=541$ for \texttt{gate\_ALL} vs.\ \texttt{none} on \textsc{Adult}; $N=540$ on \textsc{GMSC}), reporting two-sided Wilcoxon signed-rank $p$-values and Holm-adjusted $p$-values ($p_{\text{Holm}}$).
\end{itemize}

\section{Results and Ablations}
\subsection{Main Benchmark Evaluation on \textsc{Adult} and \textsc{GMSC}}
Table~\ref{tab:main_results} reports feasibility rates and conditional means across successful counterfactuals for \textsc{Adult} and \textsc{GMSC}. On \textsc{Adult}, \texttt{gate\_ALL} lowers conditional held-out \texttt{flag\_OCSVM} from $0.1625$ (\texttt{none}) to $0.0664$ and \texttt{flag\_GMM} from $0.0670$ to $0.0203$---bringing both held-out flag rates below the real favourable test reference ($0.074$ and $0.079$)---while maintaining high \texttt{robust} ($0.9151$ vs.\ $0.9012$) and reducing \texttt{found} from $0.5970$ to $0.5420$. On matched pairs ($N=541$) on \textsc{Adult}, \texttt{gate\_ALL} achieves a statistically significant reduction in held-out \textsc{OCSVM} flags ($\Delta \text{flag}_{\text{OCSVM}} = -0.0296$, $p_{\text{Holm}} = 4.0\times 10^{-4}$) with a modest paired cost increase ($\Delta \text{cost} = +0.0536$).

On \textsc{GMSC}, where unconstrained search (\texttt{none}) yields elevated conditional OOD flag rates ($\text{flag}_{\text{GMM}} = 0.5643$, $\text{flag}_{\text{OCSVM}} = 0.1637$), \texttt{gate\_ALL} reduces conditional \texttt{flag\_OCSVM} to $0.0082$ and \texttt{flag\_GMM} to $0.1265$, while increasing conditional \texttt{cost} from $0.4361$ to $0.7529$ and lowering \texttt{found} from $0.9240$ to $0.3182$ as implausible candidates are filtered out. Note that under \texttt{gate\_ALL} on \textsc{GMSC}, the conditional \texttt{IF} column ($0.3000$) reflects post-hoc evaluation relative to the reference threshold across retained trajectories and confirms substantial suppression compared to \texttt{none} ($0.8342$).

\begin{table}[H]
\centering
\caption{Feasibility rate (\texttt{found}) and conditional mean metrics across successful counterfactuals (\texttt{found} $= 1$) on \textsc{Adult} and \textsc{GMSC}.}
\label{tab:main_results}
\scriptsize
\begin{tabular}{llcccccccccc}
\toprule
Dataset & Gate & \texttt{found} & \texttt{prob} & \texttt{cost} & \texttt{n\_chg} & \texttt{robust} & \texttt{IF} & \texttt{LOF} & \texttt{kNN} & \texttt{GMM} & \texttt{OCSVM} \\
\midrule
\textsc{Adult} & \texttt{none}      & 0.5970 & 0.6246 & 2.8594 & 2.3819 & 0.9012 & 0.0938 & 0.0854 & 0.1357 & 0.0670 & 0.1625 \\
\textsc{Adult} & \texttt{gate\_IF}  & 0.5650 & 0.6253 & 2.7996 & 2.3646 & 0.9070 & 0.0195 & 0.0885 & 0.0779 & 0.0354 & 0.1044 \\
\textsc{Adult} & \texttt{gate\_LOF} & 0.5940 & 0.6253 & 2.8624 & 2.4074 & 0.9020 & 0.0875 & 0.0236 & 0.1212 & 0.0623 & 0.1481 \\
\textsc{Adult} & \texttt{gate\_ALL} & 0.5420 & 0.6259 & 2.7185 & 2.3598 & 0.9151 & 0.0037 & 0.0185 & 0.0203 & 0.0203 & 0.0664 \\
\midrule
\textsc{GMSC}  & \texttt{none}      & 0.9240 & 0.6602 & 0.4361 & 3.0169 & ---    & 0.8342 & 0.4715 & 0.4181 & 0.5643 & 0.1637 \\
\textsc{GMSC}  & \texttt{gate\_IF}  & 0.3383 & 0.7000 & 0.6915 & 3.0921 & ---    & 0.3301 & 0.3167 & 0.1094 & 0.1804 & 0.0384 \\
\textsc{GMSC}  & \texttt{gate\_LOF} & 0.8922 & 0.6616 & 0.7806 & 3.0102 & ---    & 0.8319 & 0.0408 & 0.2933 & 0.5015 & 0.1361 \\
\textsc{GMSC}  & \texttt{gate\_ALL} & 0.3182 & 0.7018 & 0.7529 & 2.9939 & ---    & 0.3000 & 0.0429 & 0.0061 & 0.1265 & 0.0082 \\
\bottomrule
\end{tabular}
\end{table}

\begin{figure}[H]
\centering
\includegraphics[width=0.82\linewidth,keepaspectratio]{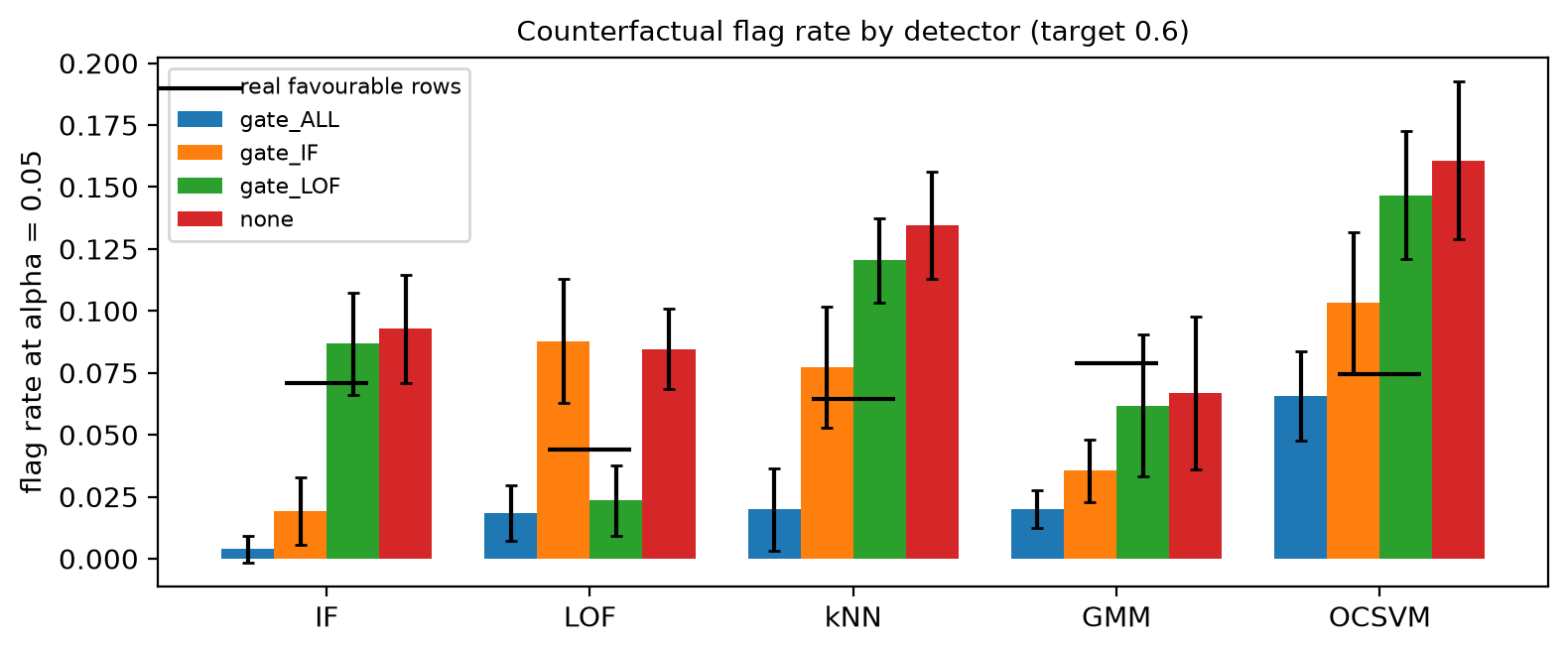}
\caption{Comparison of OOD flag rates across gates relative to real favourable reference baselines.}
\label{fig:gate_flags}
\end{figure}

\begin{figure}[H]
\centering
\includegraphics[width=0.82\linewidth,keepaspectratio]{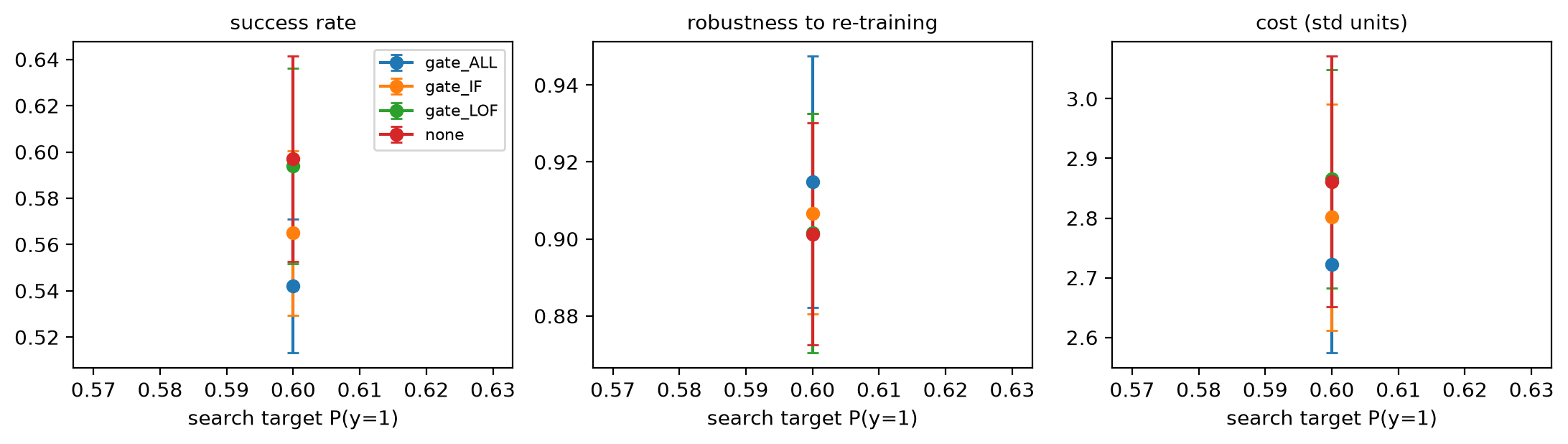}
\caption{Trade-off behavior across gates between feasibility, cost, and distributional plausibility.}
\label{fig:gate_tradeoff}
\end{figure}

\begin{figure}[H]
\centering
\includegraphics[width=0.82\linewidth,keepaspectratio]{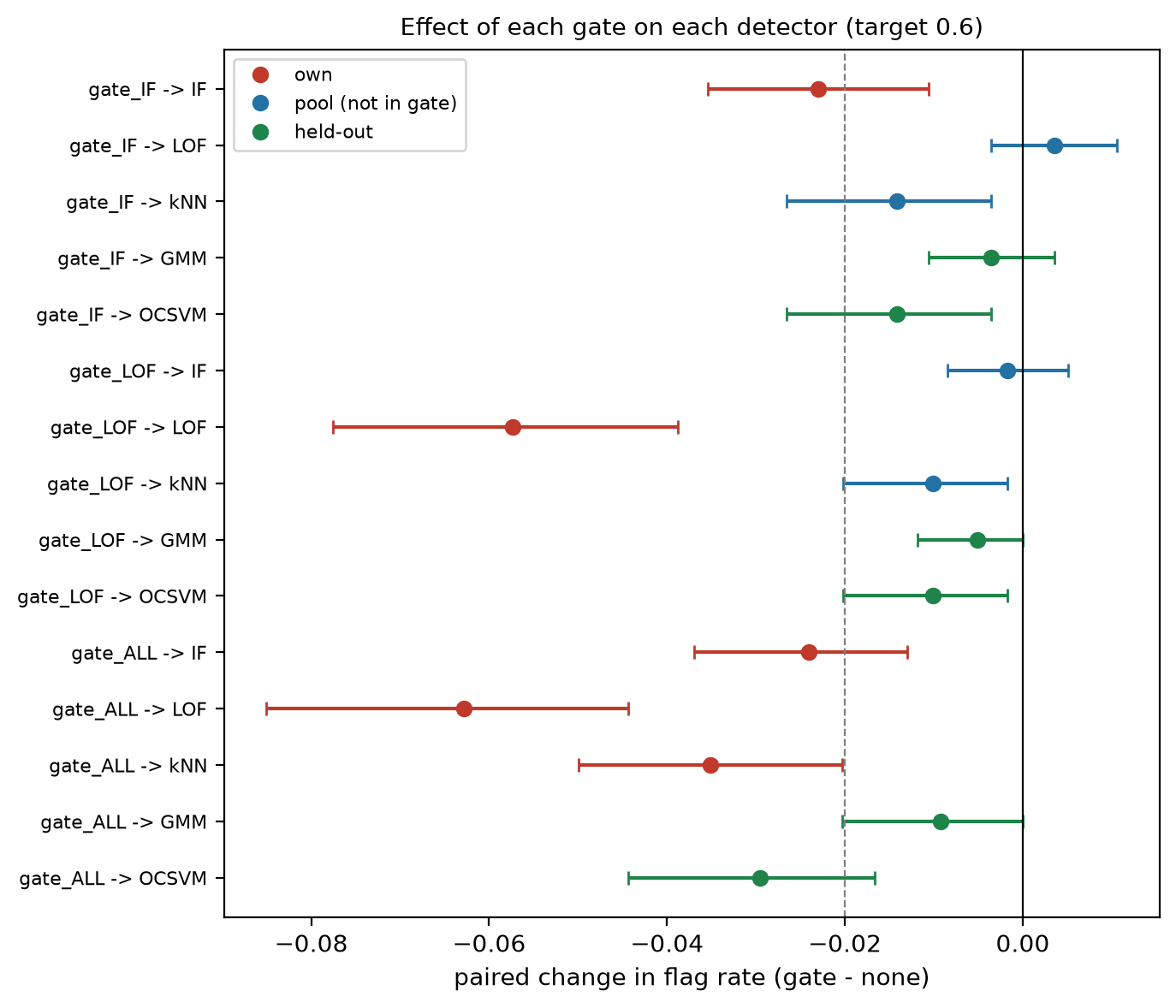}
\caption{Paired effect sizes of conformal gates relative to unconstrained recourse (\texttt{none}) on matched recourse pairs.}
\label{fig:gate_paired}
\end{figure}

Evaluating strictly on the $N=540$ matched recourse pairs on \textsc{GMSC} (where both \texttt{gate\_ALL} and \texttt{none} succeed, thus isolating within-query shifts from conditional sample composition), paired Wilcoxon signed-rank tests confirm statistically significant differences: $\Delta \text{flag}_{\text{OCSVM}} = -0.0346$ ($p = 1.0751\times 10^{-4}$), $\Delta \text{flag}_{\text{GMM}} = -0.1663$ ($p = 3.5679\times 10^{-16}$), and $\Delta \text{cost} = +0.3035$ ($p = 1.3365\times 10^{-5}$).

\subsection{Ablation Analysis (\texttt{gen10}, \texttt{gen100}, and \texttt{relax})}
Table~\ref{tab:ablations} presents ablation results on \textsc{Adult} ($\tau_p = 0.6$) comparing \texttt{gate\_IF} against \texttt{none} across generation budgets (\texttt{gen10}, \texttt{gen100}) and constraint relaxation (\texttt{relax}). Relaxing constraints (\texttt{relax}) increases \texttt{found} to $0.7233$ (\texttt{gate\_IF}) and $0.7367$ (\texttt{none}) alongside higher \texttt{n\_chg} ($\approx 2.93$), while \texttt{gate\_IF} consistently reduces \texttt{flag\_IF} across all settings (achieving a Holm-significant paired reduction $\Delta \text{flag}_{\text{IF}} = -0.0415$, $p_{\text{Holm}} = 0.0117$ across $N=217$ matched pairs under \texttt{relax}).

\begin{table}[H]
\centering
\caption{Ablation results on \textsc{Adult} ($\tau_p = 0.6$) comparing \texttt{gate\_IF} and \texttt{none} across \texttt{gen10}, \texttt{gen100}, and \texttt{relax}.}
\label{tab:ablations}
\footnotesize
\begin{tabular}{llcccccc}
\toprule
Variant & Gate & \texttt{found} & \texttt{prob} & \texttt{cost} & \texttt{n\_chg} & \texttt{flag\_IF} & \texttt{robust} \\
\midrule
\texttt{gen10}  & \texttt{gate\_IF} & 0.4800 & 0.6296 & 2.8885 & 2.3472 & 0.0208 & 0.9097 \\
\texttt{gen10}  & \texttt{none}     & 0.5067 & 0.6283 & 2.9738 & 2.3618 & 0.0987 & 0.9035 \\
\texttt{gen100} & \texttt{gate\_IF} & 0.4933 & 0.6270 & 2.9763 & 2.3851 & 0.0270 & 0.9009 \\
\texttt{gen100} & \texttt{none}     & 0.5167 & 0.6261 & 3.0115 & 2.4065 & 0.0968 & 0.9032 \\
\texttt{relax}  & \texttt{gate\_IF} & 0.7233 & 0.6326 & 2.9935 & 2.9355 & 0.0092 & 0.9217 \\
\texttt{relax}  & \texttt{none}     & 0.7367 & 0.6317 & 2.9970 & 2.9231 & 0.0679 & 0.9246 \\
\bottomrule
\end{tabular}
\end{table}

\section{Limitations and Conclusion}
This evaluation covers \textsc{Adult} and \textsc{GMSC} with \textsc{GBDT} classifiers and selected density, neighborhood, and generative/margin anomaly detectors calibrated at $\delta = 0.05$. Because conformal calibration guarantees marginal false-alarm control per detector under exchangeability rather than joint coverage across conjunctive gates, strict filtering reduces recourse feasibility on skewed credit distributions such as \textsc{GMSC}. Future extensions can investigate multi-seed parity on \textsc{GMSC}, broader recourse baselines, and structural or gradient non-conformity signals~\cite{nero,medix,tood}. Overall, conformalized OOD gating provides a statistically grounded audit against Goodhart vulnerabilities in counterfactual recourse.

\end{document}